\documentclass[11pt,a4paper]{article}
\usepackage[hyperref]{AACL-IJCNLP2020}
\usepackage{times}
\usepackage{latexsym}
\usepackage{graphicx}
\usepackage[export]{adjustbox}
\usepackage{fancyvrb}
\usepackage{listings}
\usepackage{enumitem, kantlipsum}
\usepackage{url}
\usepackage{balance}

\usepackage{amsmath}
\usepackage{subcaption}
\usepackage{amsfonts}
\usepackage{tabulary,booktabs}
\usepackage{multirow}

\usepackage{microtype}

\newcommand{\tool}{\textsc{AutoNLU}}

\aclfinalcopy 
\def\aclpaperid{4} 

\title{\tool: An On-demand Cloud-based Natural Language Understanding System for Enterprises}

\author{Nham Le \textsuperscript{1,3} \thanks{\; Equal contributions. The work was conducted while the first two authors interned at Adobe Research.} \qquad Tuan Manh Lai \textsuperscript{2,3}  \footnotemark[1] \qquad  \textbf{Trung Bui \textsuperscript{3}}  \qquad  \textbf{Doo Soon Kim \textsuperscript{3}} \\
        \textsuperscript{1} University of Waterloo, Ontario, Canada\\
	    \textsuperscript{2} University of Illinois at Urbana-Champaign, USA\\
	    \textsuperscript{3} Adobe Research, San Jose, USA
	    }

\date{}

\begin{document}
\maketitle
\begin{abstract}
With the renaissance of deep learning, neural networks have achieved promising results on many natural language understanding (NLU) tasks. Even though the source codes of many neural network models are publicly available, there is still a large gap from open-sourced models to solving real-world problems in enterprises. Therefore, to fill this gap, we introduce \tool, an on-demand cloud-based system with an easy-to-use interface that covers all common use-cases and steps in developing an NLU model. \tool\,has supported many product teams within Adobe with different use-cases and datasets, quickly delivering them working models. To demonstrate the effectiveness of \tool, we present two case studies. i) We build a practical NLU model for handling various image-editing requests in Photoshop. ii) We build powerful keyphrase extraction models that achieve state-of-the-art results on two public benchmarks. In both cases, end users only need to write a small amount of code to convert their datasets into a common format used by \tool.
\end{abstract}
\section{Introduction}
In recent years, many deep learning methods have achieved impressive results on a wide range of tasks, ranging from question answering \cite{Seo2017BidirectionalAF,lai2018review} to named entity recognition (NER) \cite{lin2019reliability,jiang-etal-2019-improved} to intent detection and slot filling \cite{wangetal2018bi,Chen2019BERTFJ}. Even though the source codes of many models are publicly available, going from an open-sourced implementation of a model for a public dataset to a production-ready model for an in-house dataset is not a simple task. Furthermore, in an enterprise, only few engineers are familiar with deep learning research and frameworks. Therefore, to facilitate the development and adoption of deep learning models within Adobe, we introduce a new system named \tool. It is an on-demand cloud-based system that enables multiple users to create and edit datasets and to train and test different state-of-the-art NLU models. \tool's main principles are:
\begin{itemize}[nosep]
    \item \textbf{Ease of use}. \tool\,aims to help users with limited technical knowledge to train and test models on their datasets. We provide GUI modules to accommodate the most common use-cases, from creating/cleaning a dataset to training/evaluating/debugging a model.
    \item \textbf{State-of-the-art models}. Users should not sacrifice performance for ease-of-use. Our built-in models provide state-of-the-art performance on multiple public datasets. \tool\,also supports  hyperparameter tuning using grid search, allowing users to fine-tune the models even further.
    \item \textbf{Scalability}. \tool\,aims to be deployed in enterprises where computing costs could be a limiting factor. We provide an on-demand architecture so that the system could be utilized as much as possible.
\end{itemize}

At Adobe, \tool\,has been used to train NLU models for different product teams, ranging from Photoshop to Document Cloud. To demonstrate the effectiveness of \tool, we present two case studies. i) We build a practical NLU model for handling various image-editing requests in Photoshop. ii) We build powerful keyphrase extraction models that achieve state-of-the-art results on two public benchmarks. In both cases, end users only need to write a small amount of code to convert their datasets into a common format used by \tool.
\section{Related work}
Closely related branches of work to ours are toolkits and frameworks designed to provide a suite of state-of-the-art NLP models to users \cite{Gong2019NeuronBlocksB,akbiketal2019flair, wang2019jiant,Zhu2020ConvLab2AO,Qi2020StanzaAP}. However, several of these works do not have a user-friendly interface. For example, \texttt{Flair} \cite{akbiketal2019flair}, \texttt{NeuronBlocks} \cite{Gong2019NeuronBlocksB}, and \texttt{jiant} \cite{wang2019jiant} require users to work with command-line interfaces. Different from these works, an end-user with no programming skill can still create powerful NLU models using our system. Furthermore, most previous works are not explicitly designed for enterprise settings where use-cases and business needs can be vastly different from team to team. On the other hand, since \tool\, is an on-demand cloud-based system, it provides more flexibility to end users.

In 2018, Google introduced AutoML Natural Language\footnote{\url{https://cloud.google.com/natural-language}}, a platform that enables users to build and deploy machine learning models for various NLP tasks. Our system is different from AutoML in the following aspects. First, AutoML uses neural architecture search (NAS) \cite{Elsken2019NeuralAS} to find the best model for the task of interest. As users are not allowed to simply choose an existing architecture, the process can be time-consuming even for simple tasks (e.g., 2$\sim$3 hours). On the other hand, \tool\,provides a rich gallery of existing architectures for NLU. In future work, we are also planning to integrate NAS into \tool. Second, as a self-hosted solution, \tool\,provides product teams of Adobe with total control over their datasets and trained models. This enhances privacy and provides more flexibility at the same time. For example, as of writing, there is no way to download a trained model from AutoML to a local machine to use it for a subsequent task. \tool\ supports it out-of-the-box.
\section{\tool}
\subsection{Components and architecture}
\begin{figure}[!t]
    \centering
    \includegraphics[width=0.4\textwidth]{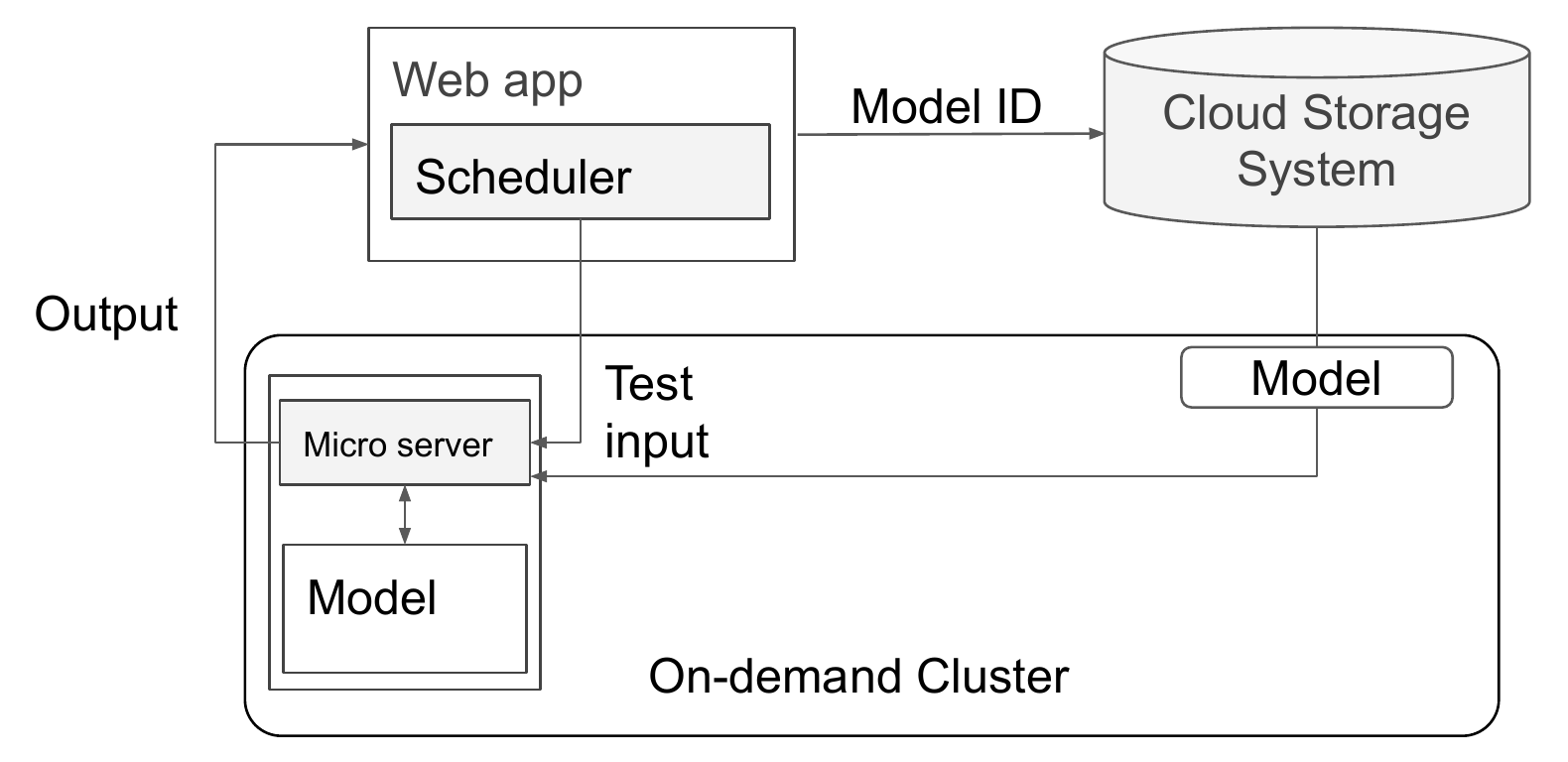}
    \caption{\tool\,architecture. In the figure is the dataflow when the user calls to the \texttt{/test} endpoint.}
    \label{fig:arc}
\end{figure}

Figure \ref{fig:arc} shows the overall architecture of our system. There are 3 main components:
\begin{itemize}[nosep]
    \item \textbf{A web application} that serves as the frontend to the users. The most important component of the application is a Scheduler that monitors the status of the cluster, then assigns jobs to the most appropriate instances, as well as spawns more/shuts off instances based on the workload to minimize the computing costs. The user interface is discussed in more detail in Section \ref{ssec:ui}.
    \item \textbf{A cloud storage system} that stores datasets, large pre-trained language models (e.g., BERT \cite{bert}), trained NLU models, and models' metadata. We use Amazon S3 as our storage system, due to its versioning support and data transfer speed to EC2 instances.
    \item \textbf{An on-demand cluster} that performs the actual training and testing. While the Lambda computing model seems to be a better fit at first thought, after careful consideration, we choose EC2 instances to prioritize user experience over some costs: in our setting, we have multiple concurrent users with small to medium datasets. If the training itself takes only 10 minutes, any amount of wait time is significant. By maintaining a certain number of always-on instances, users will always have instant interaction with the system without any delay. Cluster's instances are initiated using prebuilt images, which we discuss in Section \ref{ssec:docker image}.
\end{itemize}
  \subsection{Instance image}
\label{ssec:docker image}
Regardless of the underlying model, in each prebuilt image, an included webserver is configured to serve the following endpoints: 
\begin{itemize}[nosep]
    \item \texttt{/train} that connects to the training code of the underlying model.
    \item \texttt{/is\_free} that returns various information about the utilization of the instance (e.g, GPU memory usage).
    \item \texttt{/test} that connects to the testing code of the underlying model.
    \item \texttt{/notebook} that connects to the Jupyter Lab notebook's URL packaged in the image. 
\end{itemize}
Each image also exposes an SSH connection, authenticated using LDAP. Experienced users can also make use of the packaged TensorBoard to monitor the training process.

  \subsection{User Interface} \label{ssec:ui}
\begin{figure}[t!]
    \centering
    \includegraphics[width=0.4\textwidth,frame]{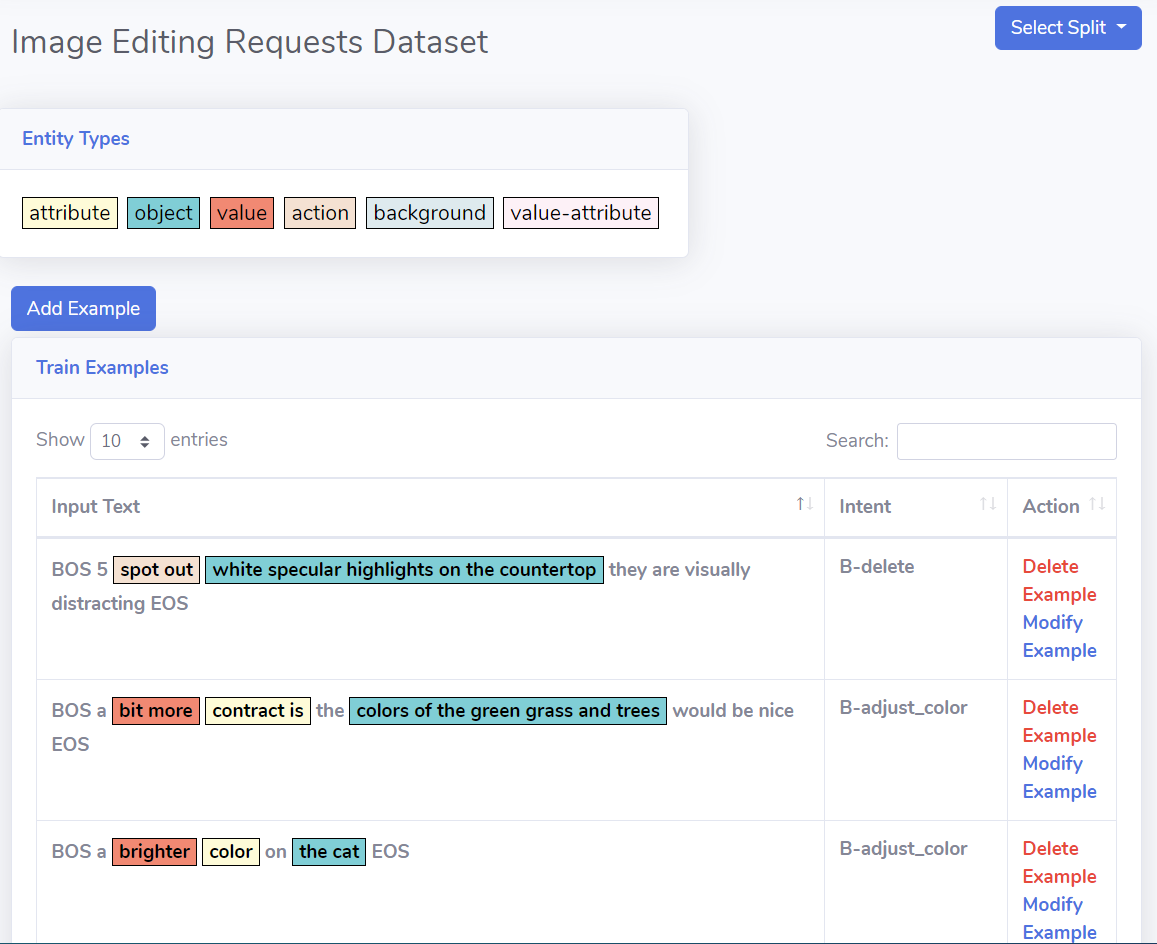}
    \caption{Dataset view of \tool.}
    \label{fig:dataset_view}
\end{figure}

\subsubsection{Dataset Tool}
Public and internal datasets come in many different formats, as they may have been collected for many years and annotated in different ways. To mitigate that, we develop an intermediate representation (IR) that is suitable for many NLU tasks and write frontends to convert common dataset formats to said IR. We also provide a converter that converts this IR back into other dataset formats, making converting a dataset from one format to another trivial. In our setting (an enterprise environment), a dataset frontend converter is the only part that may need to be written by an end-user, and we believe that it is significantly simpler than building the whole NLU pipeline.

\begin{figure}[t!]
    \centering
    \includegraphics[width=0.4\textwidth,frame]{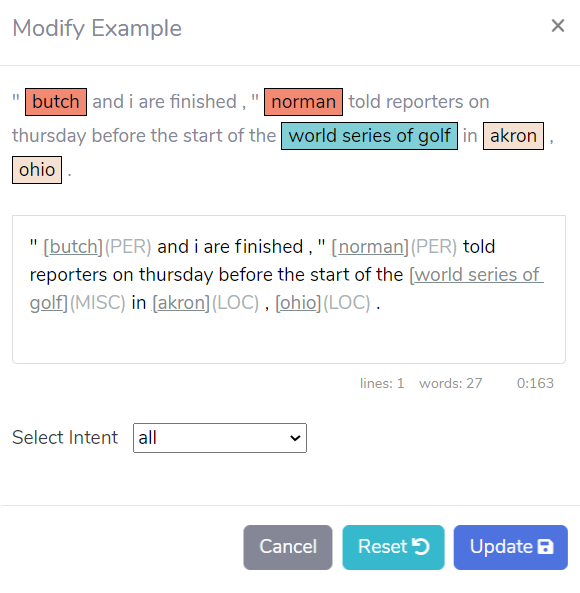}
    \caption{Edit/Add a datapoint.}
    \label{fig:edit_dp_view}
\end{figure}

Figure \ref{fig:dataset_view} shows the dataset view. Visualizing and editing datapoints are straightforward, and do not depend on the source/target dataset format (Figure \ref{fig:edit_dp_view}). While it is not common to edit a public dataset, the same is typically not true for internal datasets. Internal datasets may need to be modified and expanded based on business needs and use-cases. 


\subsubsection{Analysis Tool}
\begin{figure}[t!]
    \centering
    \includegraphics[width=0.4\textwidth, frame]{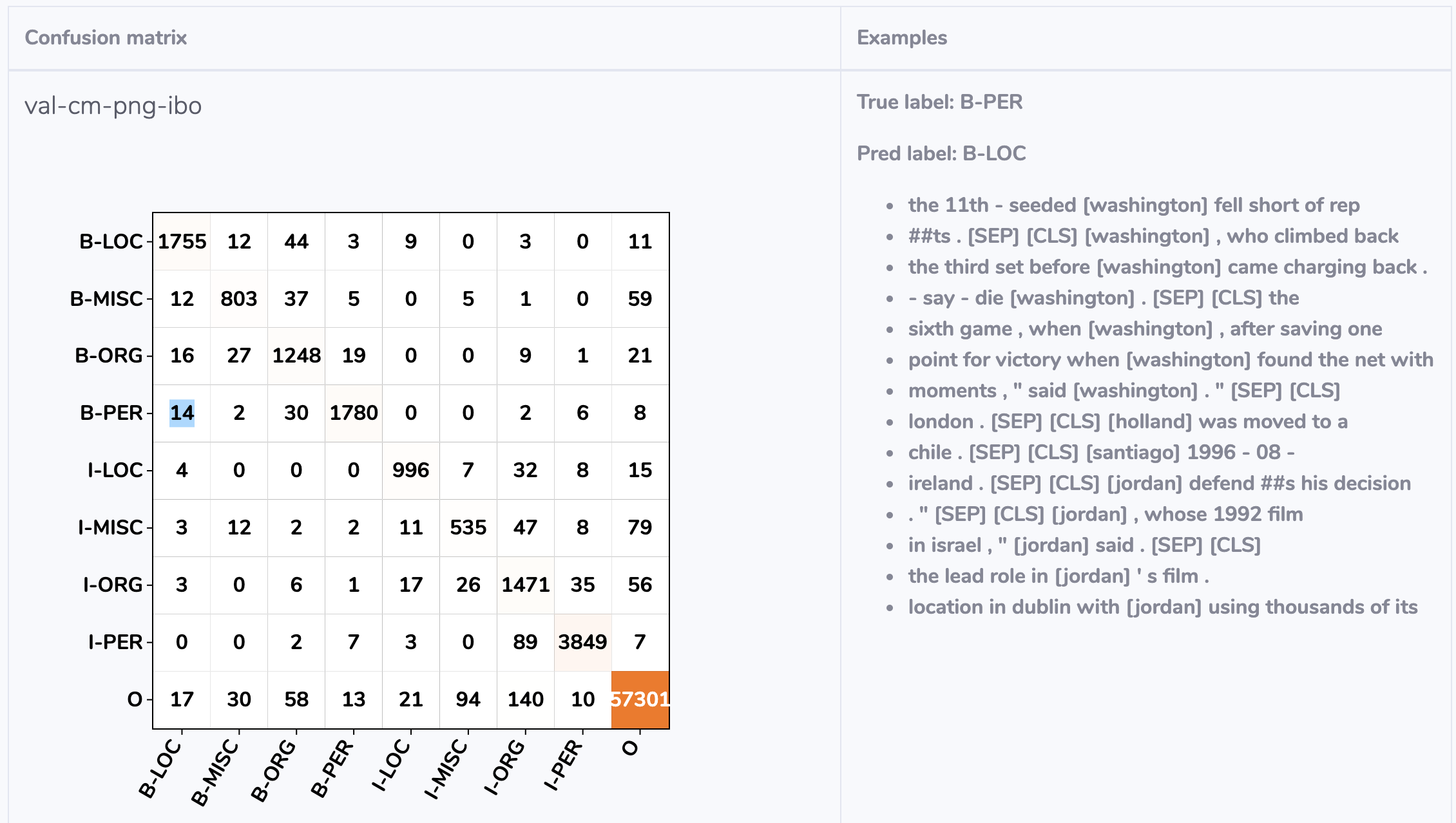}
    \caption{An example interactive confusion matrix.}
    \label{fig:int_cm}
\end{figure}

We include TensorBoard in our prebuilt images to display common training metrics. However, since our main users are typically product teams with limited experience in machine learning, we also develop interactive views to analyze the trained results. For example, Figure \ref{fig:int_cm} shows our interactive confusion matrix view: rather than just knowing that there are 14 instances in which a mention with the label ``Person'' is misclassified as ``Location'', users can click on a cell in the matrix to see which instances are misclassified. This is even more important for internal datasets: the errors may actually be in the dataset instead of the model, and we can catch it using this view. In fact, as we will demonstrate in Section \ref{ssec:case_study_1}, we have caught many labeling errors in our internal datasets using this tool.

\subsubsection{Resource Management Tool}
In most use-cases, \tool\,automatically handles resource management for the users. However, if an advanced user wants to manually manage instances' life cycle, assign a task to a specific instance, or to debug an instance, we provide a GUI to do so as well. Concretely, we provide the following functionalities:
\begin{itemize}[nosep]
    \item \textit{Create an instance with a desired hardware configuration and docker image}. By default, \tool\,creates an instance with 4 CPU cores, 8 GBs of RAM, and 1 NVIDIA V100 GPU, which are all configurable to the user's desire. The default docker image is the one containing all the supported models, but users can choose from one of the prebuilt images that contains just a single model if that's their use-case.
    \item \textit{Assign a task to an instance}. During training and testing, users can choose whether to let \tool\,to distribute the task or to assign the task to a specific instance: it is common for a product team to reserve a few instances for themselves and want to use just those instances.
    \item \textit{Access an instance's shell and files}. Since Ease-of-use is one of our core design principles, we package in all of our prebuilt images a Jupyter Lab server, with the intention of using it as a lightweight IDE/shell environment. While we also expose SSH connection to each instance, we expect users to find the Jupyter Lab a more friendly approach.
\end{itemize}
\section{Case studies}
\subsection{NLU Models for Image-Editing Requests}
\label{ssec:case_study_1}
One of the first clients of \tool\ was the Photoshop team, as we want to build a chatbot using their image-editing requests dataset \cite{manuvinakurikeetal2018edit,brixey2018system}. The dataset was collected in many years, annotated both using Amazon Mechanical Turk and by our in-house annotators. Cleaning this dataset is a challenge in itself, and in this case study, we aim to create an effective workflow to train a state-of-the-art model and clean the dataset at the same time.

We first convert the dataset into our IR, and train a simple model using the fastest algorithm provided by \tool. This initial model provides us with a rough confusion matrix, and we manually inspect cells with the biggest values. Those cells give us an insight into some systematic labeling errors, such as in Figure \ref{fig:case_study_1}. We then fix those labeling errors, either by using the dataset interface in \tool\ , or by writing scripts. With this new dataset, we retrain another model and repeat the process.

Once the fast model performance is comparable to its performance on some public datasets, such as ATIS \cite{atis}, we switch to train and fine-tune a bigger model. More specifically, we employ a joint intent classification and slot filling model based on BERT \cite{Chen2019BERTFJ}, which is already implemented in \tool. By the end of this process, we end up with a powerful NLU model, as reported in Table \ref{tab:ier}, and a cleaned dataset that is useful for subsequent tasks. The NLU model created using \tool\, outperforms a competing model created using RASA \cite{Bocklisch2017RasaOS} and a joint model of intent determination and slot filling (JIS) \cite{joint_intent_slot16} by a large margin.

\begin{figure}
\begin{lstlisting}
True label: B-adjust_brightness
Pred label: B-adjust_color
[[CLS] light ##en the vegetables [SEP]]
[[CLS] make the dirt darker in brown color [SEP]]
\end{lstlisting}
\caption{2 labeling errors captured by the interactive confusion matrix near the end of the training-cleaning process. The \#\# is the artifact from BERT tokenizer.}
\label{fig:case_study_1}
\end{figure}

\begin{table}[!t]
\centering
\begin{tabular}{|c|l|c|c|c|c|}
\hline
\multicolumn{2}{|c|}{\multirow{2}{*}{Model}} & \multicolumn{4}{c|}{Metrics}        \\ \cline{3-6} 
\multicolumn{2}{|c|}{} & \multicolumn{1}{l|}{Intent} & \multicolumn{1}{l|}{SP} & \multicolumn{1}{l|}{SR}& \multicolumn{1}{l|}{SF1} \\ \hline
\multicolumn{2}{|c|}{JIS \shortcite{joint_intent_slot16}} & 0.832 & 0.850 & 0.726 & 0.783 \\ \hline
\multicolumn{2}{|c|}{RASA} & 0.924 & 0.833 & 0.605 & 0.701 \\ \hline
\multicolumn{2}{|c|}{\tool} & \textbf{0.954} & \textbf{0.869} & \textbf{0.854} & \textbf{0.862} \\ \hline
\end{tabular}
\caption{Results on the image-editing requests dataset. Intent accuracy, slot precision, slot recall, and slot F1 scores are reported. Scores of our models are averaged over three random seeds.}
\label{tab:ier}
\end{table}
\subsection{Keyphrase Extraction Models}
Keyphrase extraction is the task of automatically extracting a small set of phrases that best describe a document. As keyphrases provide a high-level summarization of the considered document and they give the reader some clues about its contents, keyphrase extraction is a problem of great interest to the Document Cloud team of Adobe. In this case study, we aim to develop an effective keyphrase extraction system for the team.

Similar to recent works on keyphrase extraction \cite{sahrawat2020keyphrase}, we formulate the task as a sequence labeling task. Given an input sequence of tokens $\textbf{x} = \{x_1, x_2, ..., x_n\}$, the goal is to predict a sequence of labels $\textbf{y} = \{y_1, y_2, ..., y_n\}$ where $y_i \in \{\texttt{B}, \texttt{I}, \texttt{O}\}$. Here, label \texttt{B} denotes the beginning of a keyphrase, \texttt{I} denotes the continuation of a keyphrase, and \texttt{O} corresponds to tokens that are not part of any keyphrase. This formulation is naturally supported by our platform, as the task of slot filling in NLU is basically a sequence labeling task. We first collect two public datasets for keyphrase extraction: Inspec \cite{Hulth2003ImprovedAK} and SE-2017 \cite{augensteinetal2017semeval}. We then convert them to the common intermediate representation. After that, we simply use \tool\,to train and tune models. We employ the BiLSTM-CRF architecture \cite{huang2015bidirectional} that is already available in \tool. We experiment with two different pre-trained language models as the first embedding layer: BERT \cite{bert} and SciBERT \cite{Beltagy2019SciBERTAP}. Table \ref{tab:keyphrase_extraction_results} shows the results on the datasets. We see that both models created using \tool\,outperform previous models for the task, achieving new state-of-the-art results. As \tool\, can automatically perform hyperparameter tuning using grid search, models produced by \tool\, typically have satisfying performance (assuming that the selected underlying architecture is expressive enough). It is worth noting that during this entire process, the only code we need to write is for converting the Inspec and SE-2017 datasets to the IR.

\begin{table}[!t]
\centering
\begin{tabular}{|c|l|c|c|}
\hline
\multicolumn{2}{|c|}{\multirow{2}{*}{Model}} & \multicolumn{2}{c|}{Datasets}        \\ \cline{3-4} 
\multicolumn{2}{|c|}{} & \multicolumn{1}{l|}{Inspec} & \multicolumn{1}{l|}{SE-2017} \\ \hline
\multicolumn{2}{|c|}{KEA \shortcite{witten2005kea}} & 0.137 & 0.129 \\ \hline
\multicolumn{2}{|c|}{TextRank \shortcite{Mihalcea2004TextRankBO}} & 0.122 & 0.157 \\ \hline
\multicolumn{2}{|c|}{SingeRank \shortcite{Wan2008SingleDK}} & 0.123 & 0.155 \\ \hline
\multicolumn{2}{|c|}{SGRank \shortcite{Danesh2015SGRankCS}} & 0.271 & 0.211  \\ \hline
\multicolumn{2}{|c|}{Transformer \shortcite{sahrawat2020keyphrase}} & 0.595 & 0.522 \\ \hline
\multicolumn{2}{|c|}{BERT (\tool)} & 0.596 & 0.537 \\ \hline
\multicolumn{2}{|c|}{SciBERT (\tool)} & \textbf{0.598}  & \textbf{0.544} \\ \hline
\end{tabular}
\caption{Results on Inspec and SE-2017 datasets. F1 scores are reported. Scores of our models are averaged over three random seeds.}
\label{tab:keyphrase_extraction_results}
\end{table}
\section{Conclusion}
In this work, we introduce \tool, an on-demand cloud-based platform that is easy-to-use and has enabled many product teams within Adobe to create powerful NLU models. Our design principles make it an ideal candidate for enterprises who want to have an NLU system for themselves, with minimal deep learning expertise. \tool\,'s code is in the process to be open-sourced, and we invite contributors to contribute.  In future work, we will implement more advanced features such as transfer learning, knowledge distillation and neural architecture search, which have been shown to be useful in building real-world NLP systems \cite{lai2018supervised,jiang-etal-2019-improved,lai2019gated,lai2020simple,klyuchnikov2020bench}. Furthermore, we will extend our system to have more advanced analytics features \cite{murugesan2019deepcompare}, and to better support other languages \cite{nguyen2020phobert}.
\bibliography{anthology,aacl-ijcnlp2020}

\begin{thebibliography}{33}
\expandafter\ifx\csname natexlab\endcsname\relax\def\natexlab#1{#1}\fi

\bibitem[{Akbik et~al.(2019)Akbik, Bergmann, Blythe, Rasul, Schweter, and
  Vollgraf}]{akbiketal2019flair}
A.~Akbik, T.~Bergmann, Duncan Blythe, K.~Rasul, Stefan Schweter, and Roland
  Vollgraf. 2019.
\newblock Flair: An easy-to-use framework for state-of-the-art nlp.
\newblock In \emph{NAACL-HLT}.

\bibitem[{Augenstein et~al.(2017)Augenstein, Das, Riedel, Vikraman, and
  McCallum}]{augensteinetal2017semeval}
Isabelle Augenstein, Mrinal Das, Sebastian Riedel, Lakshmi Vikraman, and Andrew
  McCallum. 2017.
\newblock \href {http://arxiv.org/abs/1704.02853} {Semeval 2017 task 10:
  Scienceie - extracting keyphrases and relations from scientific
  publications}.
\newblock \emph{CoRR}, abs/1704.02853.

\bibitem[{Beltagy et~al.(2019)Beltagy, Lo, and Cohan}]{Beltagy2019SciBERTAP}
Iz~Beltagy, Kyle Lo, and Arman Cohan. 2019.
\newblock Scibert: A pretrained language model for scientific text.
\newblock In \emph{EMNLP/IJCNLP}.

\bibitem[{Bocklisch et~al.(2017)Bocklisch, Faulkner, Pawlowski, and
  Nichol}]{Bocklisch2017RasaOS}
Tom Bocklisch, Joey Faulkner, Nick Pawlowski, and Alan Nichol. 2017.
\newblock Rasa: Open source language understanding and dialogue management.
\newblock \emph{ArXiv}, abs/1712.05181.

\bibitem[{Brixey et~al.(2018)Brixey, Manuvinakurike, Le, Lai, Chang, and
  Bui}]{brixey2018system}
Jacqueline Brixey, Ramesh Manuvinakurike, Nham Le, Tuan Lai, Walter Chang, and
  Trung Bui. 2018.
\newblock A system for automated image editing from natural language commands.
\newblock \emph{arXiv preprint arXiv:1812.01083}.

\bibitem[{Chen et~al.(2019)Chen, Zhuo, and Wang}]{Chen2019BERTFJ}
Qian Chen, Zhu Zhuo, and Wen Wang. 2019.
\newblock Bert for joint intent classification and slot filling.
\newblock \emph{ArXiv}, abs/1902.10909.

\bibitem[{Danesh et~al.(2015)Danesh, Sumner, and Martin}]{Danesh2015SGRankCS}
Soheil Danesh, Tamara Sumner, and James~H. Martin. 2015.
\newblock Sgrank: Combining statistical and graphical methods to improve the
  state of the art in unsupervised keyphrase extraction.
\newblock In \emph{*SEM@NAACL-HLT}.

\bibitem[{Devlin et~al.(2018)Devlin, Chang, Lee, and Toutanova}]{bert}
Jacob Devlin, Ming{-}Wei Chang, Kenton Lee, and Kristina Toutanova. 2018.
\newblock \href {http://arxiv.org/abs/1810.04805} {{BERT:} pre-training of deep
  bidirectional transformers for language understanding}.
\newblock \emph{CoRR}, abs/1810.04805.

\bibitem[{Elsken et~al.(2019)Elsken, Metzen, and Hutter}]{Elsken2019NeuralAS}
Thomas Elsken, Jan~Hendrik Metzen, and Frank Hutter. 2019.
\newblock Neural architecture search: A survey.
\newblock \emph{ArXiv}, abs/1808.05377.

\bibitem[{Gong et~al.(2019)Gong, Shou, Lin, Sang, Yan, Yang, and
  Jiang}]{Gong2019NeuronBlocksB}
Ming Gong, Linjun Shou, Wutao Lin, Zhijie Sang, Quanjia Yan, Ze~Yang, and Daxin
  Jiang. 2019.
\newblock Neuronblocks - building your nlp dnn models like playing lego.
\newblock \emph{ArXiv}, abs/1904.09535.

\bibitem[{Hemphill et~al.(1990)Hemphill, Godfrey, and Doddington}]{atis}
Charles~T. Hemphill, John~J. Godfrey, and George~R. Doddington. 1990.
\newblock \href {https://www.aclweb.org/anthology/H90-1021} {The {ATIS} spoken
  language systems pilot corpus}.
\newblock In \emph{Speech and Natural Language: Proceedings of a Workshop Held
  at Hidden Valley, {P}ennsylvania, June 24-27,1990}.

\bibitem[{Huang et~al.(2015)Huang, Xu, and Yu}]{huang2015bidirectional}
Zhiheng Huang, Wei Xu, and Kai Yu. 2015.
\newblock Bidirectional lstm-crf models for sequence tagging.
\newblock \emph{arXiv preprint arXiv:1508.01991}.

\bibitem[{Hulth(2003)}]{Hulth2003ImprovedAK}
Anette Hulth. 2003.
\newblock Improved automatic keyword extraction given more linguistic
  knowledge.
\newblock In \emph{EMNLP}.

\bibitem[{Jiang et~al.(2019)Jiang, Hu, Xiao, Zhang, and
  Zhu}]{jiang-etal-2019-improved}
Yufan Jiang, Chi Hu, Tong Xiao, Chunliang Zhang, and Jingbo Zhu. 2019.
\newblock Improved differentiable architecture search for language modeling and
  named entity recognition.
\newblock In \emph{EMNLP/ICJNLP}.

\bibitem[{Klyuchnikov et~al.(2020)Klyuchnikov, Trofimov, Artemova, Salnikov,
  Fedorov, and Burnaev}]{klyuchnikov2020bench}
Nikita Klyuchnikov, Ilya Trofimov, Ekaterina Artemova, Mikhail Salnikov, Maxim
  Fedorov, and Evgeny Burnaev. 2020.
\newblock Nas-bench-nlp: Neural architecture search benchmark for natural
  language processing.
\newblock \emph{arXiv preprint arXiv:2006.07116}.

\bibitem[{Lai et~al.(2018{\natexlab{a}})Lai, Bui, Lipka, and
  Li}]{lai2018supervised}
Tuan Lai, Trung Bui, Nedim Lipka, and Sheng Li. 2018{\natexlab{a}}.
\newblock Supervised transfer learning for product information question
  answering.
\newblock In \emph{2018 17th IEEE International Conference on Machine Learning
  and Applications (ICMLA)}, pages 1109--1114. IEEE.

\bibitem[{Lai et~al.(2019)Lai, Tran, Bui, and Kihara}]{lai2019gated}
Tuan Lai, Quan~Hung Tran, Trung Bui, and Daisuke Kihara. 2019.
\newblock A gated self-attention memory network for answer selection.
\newblock \emph{arXiv preprint arXiv:1909.09696}.

\bibitem[{Lai et~al.(2018{\natexlab{b}})Lai, Bui, and Li}]{lai2018review}
Tuan~Manh Lai, Trung Bui, and Sheng Li. 2018{\natexlab{b}}.
\newblock \href {https://www.aclweb.org/anthology/C18-1181} {A review on deep
  learning techniques applied to answer selection}.
\newblock In \emph{Proceedings of the 27th International Conference on
  Computational Linguistics}, pages 2132--2144, Santa Fe, New Mexico, USA.
  Association for Computational Linguistics.

\bibitem[{Lai et~al.(2020)Lai, Tran, Bui, and Kihara}]{lai2020simple}
Tuan~Manh Lai, Quan~Hung Tran, Trung Bui, and Daisuke Kihara. 2020.
\newblock A simple but effective bert model for dialog state tracking on
  resource-limited systems.
\newblock In \emph{ICASSP 2020-2020 IEEE International Conference on Acoustics,
  Speech and Signal Processing (ICASSP)}, pages 8034--8038. IEEE.

\bibitem[{Lin et~al.(2019)Lin, Liu, Ji, Yu, and Han}]{lin2019reliability}
Ying Lin, Liyuan Liu, Heng Ji, Dong Yu, and Jiawei Han. 2019.
\newblock \href {https://doi.org/10.18653/v1/P19-1016} {Reliability-aware
  dynamic feature composition for name tagging}.
\newblock In \emph{Proceedings of the 57th Annual Meeting of the Association
  for Computational Linguistics}, pages 165--174, Florence, Italy. Association
  for Computational Linguistics.

\bibitem[{Manuvinakurike et~al.(2018)Manuvinakurike, Brixey, Bui, Chang, Kim,
  Artstein, and Georgila}]{manuvinakurikeetal2018edit}
Ramesh~R. Manuvinakurike, Jacqueline Brixey, Trung Bui, W.~Chang, Doo~Soon Kim,
  Ron Artstein, and Kallirroi Georgila. 2018.
\newblock Edit me: A corpus and a framework for understanding natural language
  image editing.
\newblock In \emph{LREC}.

\bibitem[{Mihalcea and Tarau(2004)}]{Mihalcea2004TextRankBO}
Rada Mihalcea and Paul Tarau. 2004.
\newblock Textrank: Bringing order into text.
\newblock In \emph{EMNLP}.

\bibitem[{Murugesan et~al.(2019)Murugesan, Malik, Du, Koh, and
  Lai}]{murugesan2019deepcompare}
Sugeerth Murugesan, Sana Malik, Fan Du, Eunyee Koh, and Tuan~Manh Lai. 2019.
\newblock Deepcompare: Visual and interactive comparison of deep learning model
  performance.
\newblock \emph{IEEE computer graphics and applications}, 39(5):47--59.

\bibitem[{Nguyen and Nguyen(2020)}]{nguyen2020phobert}
Dat~Quoc Nguyen and Anh~Tuan Nguyen. 2020.
\newblock Phobert: Pre-trained language models for vietnamese.
\newblock \emph{arXiv preprint arXiv:2003.00744}.

\bibitem[{Qi et~al.(2020)Qi, Zhang, Zhang, Bolton, and
  Manning}]{Qi2020StanzaAP}
Peng Qi, Yuhao Zhang, Yuhui Zhang, Jason Bolton, and Christopher~D. Manning.
  2020.
\newblock Stanza: A python natural language processing toolkit for many human
  languages.
\newblock In \emph{ACL}.

\bibitem[{Sahrawat et~al.(2020)Sahrawat, Mahata, Zhang, Kulkarni, Sharma,
  Gosangi, Stent, Kumar, Shah, and Zimmermann}]{sahrawat2020keyphrase}
Dhruva Sahrawat, Debanjan Mahata, Haimin Zhang, Mayank Kulkarni, Agniv Sharma,
  Rakesh Gosangi, Amanda Stent, Yaman Kumar, Rajiv~Ratn Shah, and Roger
  Zimmermann. 2020.
\newblock Keyphrase extraction as sequence labeling using contextualized
  embeddings.
\newblock In \emph{European Conference on Information Retrieval}, pages
  328--335. Springer.

\bibitem[{Seo et~al.(2017)Seo, Kembhavi, Farhadi, and
  Hajishirzi}]{Seo2017BidirectionalAF}
Minjoon Seo, Aniruddha Kembhavi, Ali Farhadi, and Hannaneh Hajishirzi. 2017.
\newblock Bidirectional attention flow for machine comprehension.
\newblock \emph{ArXiv}, abs/1611.01603.

\bibitem[{Wan and Xiao(2008)}]{Wan2008SingleDK}
Xiaojun Wan and Jianguo Xiao. 2008.
\newblock Single document keyphrase extraction using neighborhood knowledge.
\newblock In \emph{AAAI}.

\bibitem[{Wang et~al.(2019)Wang, Tenney, Pruksachatkun, Yu, Hula, Xia,
  Pappagari, Jin, McCoy, Patel, Huang, Phang, Grave, Liu, Kim, Htut, F'{e}vry,
  Chen, Nangia, Mohananey, Kann, Bordia, Patry, Benton, Pavlick, and
  Bowman}]{wang2019jiant}
Alex Wang, Ian~F. Tenney, Yada Pruksachatkun, Katherin Yu, Jan Hula, Patrick
  Xia, Raghu Pappagari, Shuning Jin, R.~Thomas McCoy, Roma Patel, Yinghui
  Huang, Jason Phang, Edouard Grave, Haokun Liu, Najoung Kim, Phu~Mon Htut,
  Thibault F'{e}vry, Berlin Chen, Nikita Nangia, Anhad Mohananey, Katharina
  Kann, Shikha Bordia, Nicolas Patry, David Benton, Ellie Pavlick, and
  Samuel~R. Bowman. 2019.
\newblock \texttt{jiant} 1.2: A software toolkit for research on
  general-purpose text understanding models.
\newblock \url{http://jiant.info/}.

\bibitem[{Wang et~al.(2018)Wang, Shen, and Jin}]{wangetal2018bi}
Yu~Wang, Yilin Shen, and Hongxia Jin. 2018.
\newblock A bi-model based rnn semantic frame parsing model for intent
  detection and slot filling.
\newblock \emph{ArXiv}, abs/1812.10235.

\bibitem[{Witten et~al.(2005)Witten, Paynter, Frank, Gutwin, and
  Nevill-Manning}]{witten2005kea}
Ian~H Witten, Gordon~W Paynter, Eibe Frank, Carl Gutwin, and Craig~G
  Nevill-Manning. 2005.
\newblock Kea: Practical automated keyphrase extraction.
\newblock In \emph{Design and Usability of Digital Libraries: Case Studies in
  the Asia Pacific}, pages 129--152. IGI global.

\bibitem[{Zhang and Wang(2016)}]{joint_intent_slot16}
Xiaodong Zhang and Houfeng Wang. 2016.
\newblock A joint model of intent determination and slot filling for spoken
  language understanding.
\newblock In \emph{IJCAI}.

\bibitem[{Zhu et~al.(2020)Zhu, Zhang, Fang, Li, Takanobu, chao Li, Peng, Gao,
  Zhu, and Huang}]{Zhu2020ConvLab2AO}
Qi~Zhu, Zheng Zhang, Yan Fang, Xiang Li, Ryuichi Takanobu, Jin chao Li, Baolin
  Peng, Jianfeng Gao, Xiao-Yan Zhu, and Minlie Huang. 2020.
\newblock Convlab-2: An open-source toolkit for building, evaluating, and
  diagnosing dialogue systems.
\newblock \emph{ArXiv}, abs/2002.04793.

\end{thebibliography}
\bibliographystyle{acl_natbib}
\end{document}